# Utilizing Earth Foundation Models to Enhance the Simulation Performance of Hydrological Models with AlphaEarth Embeddings

**Pengfei Qu[1], Lei Ye[2], Wenyu Ouyang[2], Chi Zhang[2], Yikai Chai[2], Shuolong Xu[1], Yongri Piao[3], Miao Zhang[4], Huchuan Lu[5]**

[1] School of Computer Science and Technology, Dalian University of Technology, Dalian, China

[2] School of Infrastructure Engineering, Dalian University of Technology, Dalian, China

[3] School of Information and Communication Engineering, Dalian University of Technology, Dalian, China

[4] School of Software, Dalian University of Technology, Dalian, China

[5] School of Artificial Intelligence, Dalian University of Technology, Dalian, China

Corresponding author: Wenyu Ouyang (wenyuouyang@dlut.edu.cn)

## Key Points:

- AEF embeddings provide stronger PUB transferability than CAMELS attributes.
- Donor-selection benefits depend on static inputs and pool size, and more donors do not consistently improve predictions.
- The benefits of AEF inputs diminish when entire AEF-defined basin groups are excluded from training.

## Abstract

Accurate prediction in ungauged basins (PUB) requires effective transfer across hydrologically diverse regions. While deep learning has advanced large-sample hydrological modeling, its cross-regional generalization depends in part on the static basin descriptors used to guide model behavior. Earth Foundation Models (EFMs) provide a promising alternative by learning environmental representations from multi-modal satellite data. In this study, we evaluate the 64-dimensional AlphaEarth Foundations (AEF) embeddings as a new form of static basin representation and assess their value for streamflow prediction under both in-sample (IS) and out-of-sample (OOS) settings using the CAMELS dataset. Results show that AEF embeddings offer higher OOS robustness than CAMELS attributes, despite nearly identical IS performance. Mutual-information analysis reveals statistical associations between AEF embeddings and conventional catchment attributes related to terrain, vegetation, and rooting characteristics. We further investigate similarity-based donor-basin selection for PUB using CAMELS attributes and AEF embeddings. Across 100 target basins, AEF static inputs generally improved PUB performance. With AEF inputs, similarity-based donor selection outperformed random selection for smaller donor pools, but its advantage diminished as pool size increased. Adding more donor basins did not consistently improve performance. Leave-one-cluster-out experiments showed AEF input benefits depended on basin partitioning, suggesting that transferability remains sensitive to the catchment conditions represented during training. Overall, this study demonstrates that Earth Foundation Model embeddings provide a transferable representation for PUB prediction and underscores the importance of basin representation quality and donor pool design, motivating future research toward unified hydrological foundation models.

## Plain-language summary

Predicting river flow in places without streamflow records is challenging because basins respond differently to climate, terrain, vegetation, and soils. Traditional basin attributes describe some of these differences, but they cannot fully represent the complexity of natural environments. This study examines whether AlphaEarth Foundations embeddings, which are learned from large collections of satellite images rather than designed by experts, offer a more informative way to describe basin characteristics. These embeddings summarize patterns in vegetation, land surface properties, and long-term environmental dynamics. We find that models using them achieve higher accuracy when predicting flows in basins not used for training. We further investigate how selecting appropriate donor basins influences prediction in ungauged regions. Using these embeddings as model inputs generally improves predictions, while choosing similar rather than random basins is more helpful when fewer training basins are used. Meanwhile, adding more donor basins does not consistently improve accuracy. The range of training basins also mattered: the AlphaEarth advantage disappeared when an entire type of basin was missing from training. The results show that satellite-derived basin representations can improve prediction in ungauged basins, while reliable transfer still depends on the range of basin conditions represented during training.

## 1 Introduction

Accurate prediction in ungauged basins remains one of the most persistent challenges in hydrology. The Prediction in Ungauged Basins (PUB) initiative launched in 2003 reframed this issue as a decade-long community effort to improve process understanding, quantify uncertainties, and develop models capable of transferring information across basins with sparse or nonexistent observations (Hrachowitz et al., 2013; Sivapalan et al., 2003). A long-standing strategy for PUB is regionalization, in which model parameters or training data for an ungauged target basin are borrowed from gauged donor basins. This process relies fundamentally on the hypothesis that hydrological behaviors can be transferred between locations that share high similarity in their physical and climatic characteristics (Pool et al., 2021). Although foundational studies have demonstrated the potential of these donor-selection schemes, they also show that prediction performance is sensitive to how similarity is defined and implemented (Oudin et al., 2008).

Deep learning has reshaped the landscape of regional rainfall-runoff modeling, as Long Short-Term Memory (LSTM) models (Hochreiter and Schmidhuber, 1997) trained on large collections of basins have demonstrated strong information sharing and generalization capacity, achieving impressive performance even in ungauged settings (Fang et al., 2022; Kratzert et al., 2019a). This success has led to the prevailing view that multi-basin learning should supplant single-basin training (Kratzert et al., 2024). However, the tendency to continually enlarge the training dataset introduces an important limitation. Simply expanding the training set does not always yield better predictions for every catchment. Evidence from studies covering more than 3,000 U.S. basins indicates that pooling all basins into a single training set can be suboptimal, largely because of the substantial hydrological heterogeneity among nonreference basins affected by human activities (Ouyang et al., 2021). Critically, even within the carefully curated CAMELS dataset (Addor et al., 2017), a benchmark collection of reference basins with minimal human impact, training a regional deep learning model on all available data without proper screening can degrade performance at the local scale (Nai et al., 2024; Yu et al., 2024). Thus, identifying a small yet informative set of training basins remains an underexplored but highly influential lever for improving PUB performance.

The challenge of selecting these informative donors has traditionally relied on distance-based similarity analysis using readily quantifiable catchment descriptors (He et al., 2011). These hand-crafted attributes spanning physiographic, climatic, and geological factors, are typically evaluated individually or combined into composite metrics such as standardized Euclidean distance. However, conventional descriptors alone are often inadequate (Tarasova et al., 2024). Hydrological processes can vary sharply over space, meaning that spatial proximity only loosely reflects functional similarity. Furthermore, these static attributes are often too sparse to capture the complex, dynamic nature of catchment behavior, resulting in distance metrics that do not map effectively onto true hydrological similarity.

The integration of deep learning has introduced sophisticated approaches to overcome the limitations of traditional, hand-crafted similarity measures. Early studies demonstrated that regional LSTM models could implicitly exploit static attributes to infer streamflow generation mechanisms (Kratzert et al., 2019b). Recent developments take a more explicit approach by deriving improved similarity rules directly from data. One line of work defines similarity through the classification of streamflow generation mechanisms based on hydrological signatures, thereby grouping catchments with more homogeneous behavior (Yu et al., 2024). Another, more data-driven direction introduces dynamic basin-affinity metrics that quantify how the learning gradient from one basin affects the loss function of a target basin. This allows for the automatic identification of training subsets that positively contribute to the target model while filtering out "mutual noise" (Nai et al., 2024).

These emerging approaches represent important advances in characterizing basin similarity and identifying informative donor basins. They assess relationships among catchments through streamflow behavior or model training. A related question is whether broader land-surface observations can provide informative catchment descriptions independently of the rainfall–runoff modeling task. Conventional catchment descriptors offer an explicit physical basis for describing basin differences, but a finite set of hand-crafted attributes may not fully represent the spatial heterogeneity or the combined influence of physical processes on hydrological responses. Richer descriptions derived from Earth observation data could therefore provide a complementary basis for characterizing catchments and supporting information transfer to ungauged basins.

Earth observation (EO) foundation models (FMs) offer a promising way to derive such catchment descriptions from large-scale land-surface observations (Lacoste et al., 2023). While the deep learning approaches previously discussed are powerful, they are fundamentally constrained by their supervised training on rainfall-runoff tasks within a limited set of gauged basins. Consequently, their learned relationships remain dependent on the specific dataset and tasks, potentially limiting generalization to novel hydrological regimes. FMs offer a paradigm shift. These self-supervised models are pre-trained on massive archives of unlabeled, multi-modal geospatial data, scaling from terabytes to even petabytes, covering the entire globe (Mai et al., 2024). Critically, FMs are not trained on direct streamflow labels; instead, they learn the global intrinsic patterns and relationships within the Earth system data itself (e.g., phenology, surface moisture, land cover changes) (Kraabel et al., 2025). Through this self-supervised, global pre-training, FMs distill complex environmental signals into general-purpose "embeddings" (Lacoste et al., 2021) that describe each location with a rich, transferable representation. Notable examples include SatMAE (Cong et al., 2022), which leverages masked autoencoders (MAE) (He et al., 2022) to learn representations from temporal and multi-spectral satellite imagery, and Prithvi-EO-2.0 (Szwarcman et al., 2025), a multi-temporal Geospatial Foundation Model (GFM) trained on millions of samples from Harmonized Landsat and Sentinel-2 (HLS) data. These GFMs provide high-dimensional vectors, such as the 256-dimensional SatCLIP location embeddings (Klemmer et al., 2025), that implicitly encode a location's visual and environmental characteristics, like climate zones or land use patterns, enabling the identification of functionally similar locations even when they are geographically distant.

A prominent, publicly accessible model that operationalizes this concept is Google DeepMind's AlphaEarth Foundations (AEF) (Brown et al., 2025). Hosted on Earth Engine, AEF provides publicly accessible, 64-dimensional satellite embeddings (annual, 2017-2024) that serve as analysis-ready representations. While these embeddings lack explicit hydrologic semantics, they offer a dense, rich context of the environmental state. Consequently, AEF presents a novel and potentially powerful alternative to traditional hand-crafted attributes for characterizing catchments, opening a new avenue to enhance streamflow prediction in ungauged basins.

This study first aims to evaluate the intrinsic value of AEF embeddings by assessing their performance in direct streamflow prediction under both in-sample (IS) and out-of-sample (OOS) conditions (Heudorfer et al., 2025). By directly integrating these embeddings into a deep learning model for streamflow prediction, we can evaluate their predictive utility and assess their contribution to model performance compared with conventional basin attributes. Following this, we address the PUB setting by fixing a single target basin and constructing its training set exclusively from other basins chosen by a similarity rule. We evaluate two operational definitions of similarity: cosine similarity computed from catchment attributes and cosine similarity computed from AEF embeddings. Using a large-sample benchmark (e.g., CAMELS) to standardize model inputs and evaluation, we ask: how do donor-pool size and composition affect PUB performance, and what are the respective contributions of donor selection and static catchment representation? Crucially, this investigation includes a scaling analysis to examine how predictive performance evolves as the number of donor basins increases. This allows us to examine whether adding more donor basins consistently improves PUB performance and how this relationship varies with the static catchment representation. Finally, we compare how AEF embeddings and conventional attributes organize basins and assess model transferability when entire basin groups are excluded from training.

The remainder of the paper is organized as follows. Section 2 introduces the datasets (hydrometeorological time series, catchment attributes, and AEF embeddings), the model architecture and similarity computations. Section 3 details the experimental design for evaluating static inputs and donor-pool scaling, including target selection, donor selection under the two similarity definitions, training protocols, evaluation metrics and ablation settings. Section 4 presents comparative analysis, focusing on predictive skill and the sensitivity of performance to donor-pool size. Section 5 discusses how AEF embeddings organize basins compared with conventional attributes and examines their utility and limitations for cross-regime generalization. Finally, Section 6 concludes the study and outlines directions for future work.

## 2 Data and Methods

### 2.1 CAMELS-US dataset

We used all 671 catchments from the CAMELS dataset as experimental data. For dynamic input data we used precipitation, daylength, shortwave radiation, minimum/maximum temperature, vapor pressure and potential evapotranspiration from the Daymet (Thornton et al., 2016) data set in the CAMELS dataset. These features were subsequently fed into two model variants utilized in this study. One model variant utilized a set of 17 physiographic catchment attributes derived from the CAMELS dataset. Further details on the meteorological forcing variables and CAMELS-derived static basin attributes used as model inputs are provided in Supporting Information Text S1 and Table S1. This feature set is consistent with that employed by Feng, et al. (2020). Given the exhaustive description of these features provided in that prior work, we refrain from further discussion here. Another model variant utilized a 64-dimensional embedding vector sourced from the Satellite Embedding dataset generated by the AEF.

### 2.2 AlphaEarth Foundations Embedding

AEF derives its representations from a vast archive of global satellite imagery, utilizing sensors such as Sentinel-1/2 and Landsat to cover the Earth's land surface. Rather than training on raw time series indiscriminately, the model employs a globally balanced sampling grid. This strategy ensures that the learned features meaningfully reflect the diversity of Earth's biomes, climates, and land-surface conditions. Through unified training pipelines, AEF condenses these multi-temporal and multi-spectral inputs into compact embedding vectors. These embeddings are learned in part by training decoders to reconstruct source observations from the compressed representations, together with observation times and sensor metadata (Brown et al., 2025). These representations capture both fine spatial details and broader environmental patterns, effectively summarizing complex land-surface dynamics, including vegetation phenology and surface texture, into a low-dimensional format.

The final output consists of 64-dimensional embedding vectors generated at a 10-meter spatial resolution, spanning the eight-year period from 2017 to 2024. For a foundation model, a general usage is to use its output as the input for downstream applications (the foundation model itself is the upstream foundation). Hence, a simple usage is to replace the original static basin attributes with the 64-dimensional AEF embedding while keeping the meteorological forcing inputs unchanged. These vectors are then paired with the corresponding ground-truth labels and fed into the model for training. During the testing phase, the embedding vectors extracted from the test set are input to the trained model to facilitate the prediction of target labels. This approach effectively utilizes the highly compact and semantically rich representations learned by AEF as a foundational feature set, significantly simplifying the input feature engineering for various downstream geospatial tasks.

For each study basin, we obtained a single representative vector by averaging the pixel-level embeddings across both spatial and temporal dimensions. This aggregation provides a fixed, basin-level representation for comparison with conventional static catchment attributes. Although AEF embeddings are provided annually, the spatially averaged basin vectors showed high interannual consistency during 2017–2024, with a median cosine similarity of 0.983 between consecutive years and 0.990 between annual vectors and their eight-year temporal averages (Supporting Information Text S2 and Figure S1). The consistency observed during 2017–2024 does not establish that equivalent representations would have been obtained during the CAMELS streamflow period used here (1980–2014). Our objective, however, is to evaluate the predictive utility of AEF embeddings as static catchment descriptors, rather than to reconstruct historical catchment conditions. All model variants were trained and evaluated using the same streamflow records, meteorological forcings, temporal partitions, and basin splits, providing a consistent basis for comparing descriptor representations within this modeling setting.

## 2.3 Model and Similarity definitions

The LSTM network is a type of recurrent neural network that incorporates dedicated memory cells capable of storing information over extended time periods. Information flow within the LSTM is controlled by a specific configuration of operations known as gates. These memory cells are, in essence, analogous to the state vector within a traditional dynamical systems model, positioning LSTMs as a potentially ideal candidate for modeling complex dynamical systems. Compared to other recurrent neural network architectures, LSTMs mitigate the issues of exploding and vanishing gradients, which enables them to effectively learn long-term dependencies between input and output features. All experiments in this study utilized a simple LSTM model and multilayer perceptron that serves as a feature transformation component (Figure 1). The MLP is composed of fully connected layers with nonlinear activation functions, enabling it to bring heterogeneous inputs into a more compact and informative latent representation. This preprocessing step helps combine static basin characteristics with dynamic meteorological inputs before they enter the LSTM.

As illustrated in Figure 1, the model first integrates static basin descriptors including CAMELS attributes or AEF embeddings with dynamic meteorological inputs. These two types of information play complementary roles. The static attributes provide basin characteristics that remain constant, whereas the dynamic features capture day-to-day climatic variations that drive hydrological responses. After concatenation, the combined input vector is passed through a multilayer perceptron that maps the heterogeneous inputs into a shared latent feature space. The transformed sequence is then processed by the LSTM, which extracts temporal dependencies essential for representing runoff-generation processes. Finally, a feedforward neural network converts the LSTM outputs into streamflow predictions. This architecture allows static basin properties to modulate how meteorological forcings are interpreted, thereby enabling the model to learn basin-specific hydrologic behavior while retaining a uniform network structure across all basins.

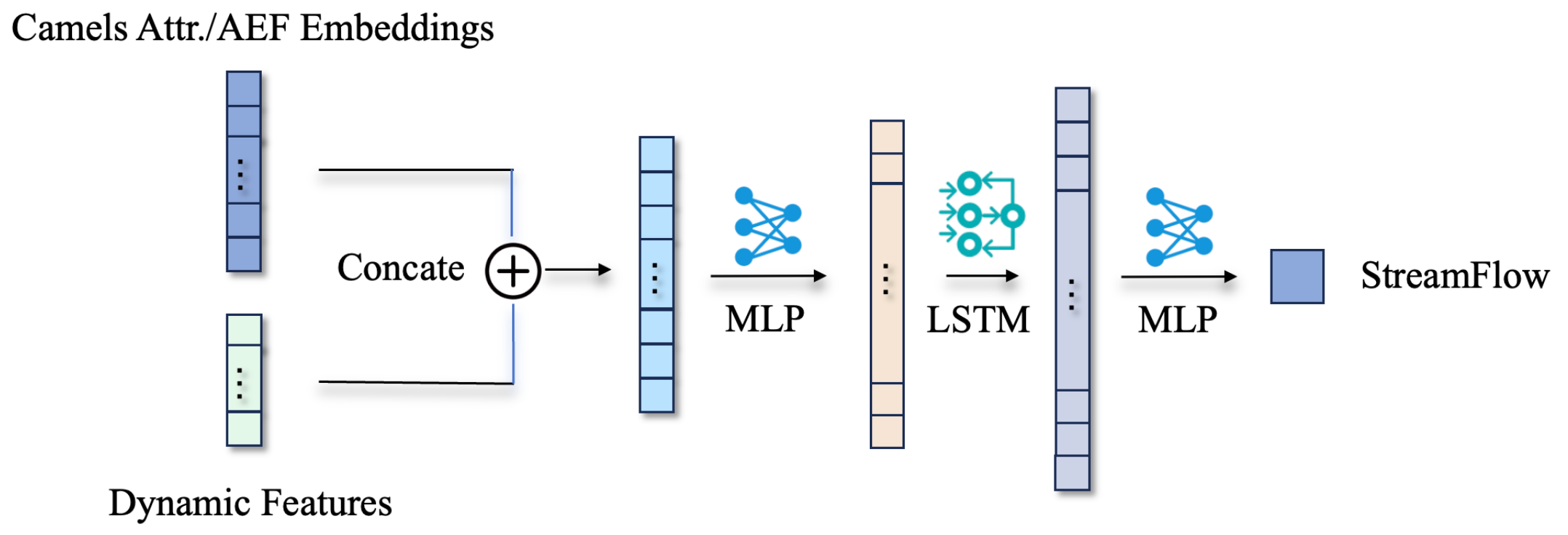

Figure 1. The model architecture used in the study.

To improve the accuracy of predictions in ungauged basins (PUB), we opted to train the LSTM model using a set of basins that resemble the target catchment. Basin similarity was evaluated using cosine similarity computed under two different representations.

(1) Attribute similarity. Construct a standardized attribute vector for each basin and compute pairwise cosine similarity.

(2) AEF embedding similarity. Use the AEF embedding vector from 2.2 and compute cosine similarity.

Formally, for a representation $z_i$ of basin $i$, the similarity to basin $j$ is

$$s(i,j) = \frac{z_i^T z_j}{\|z_i\|\|z_j\|}$$

We z-score attributes across the candidate donor set; for AEF embeddings, we apply per-dimension standardization before similarity computation. The data processing workflow is shown in Figure 2.

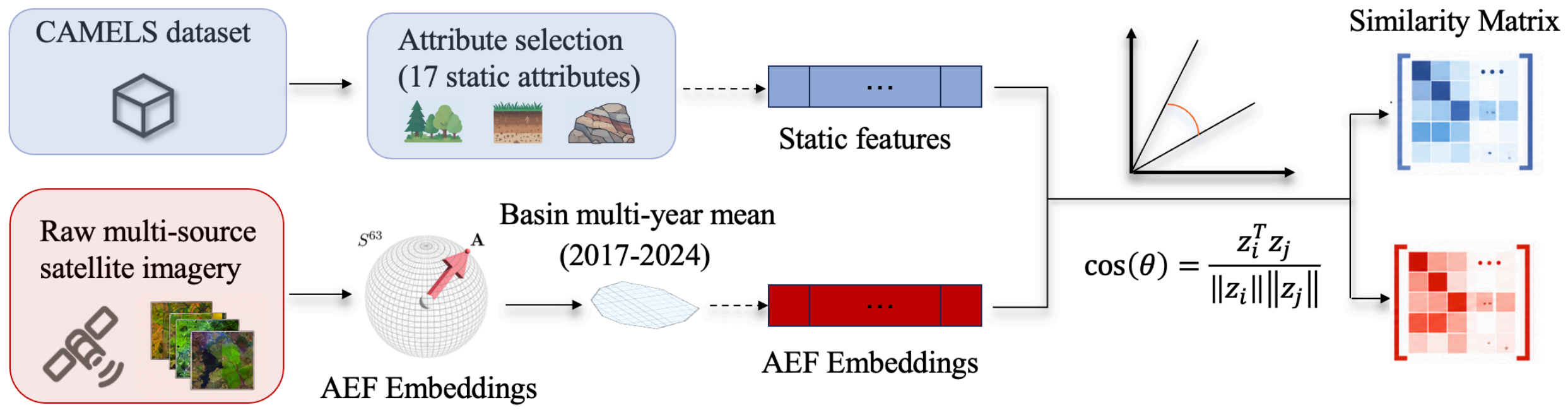

Figure 2. Two methods of calculating similarity matrices between basins.

## 3 Experiments

### Experiment A

We first examine whether AEF can improve simulation performance relative to conventional static basin attributes, thereby assessing whether AEF capture additional information that is beneficial for streamflow prediction. We used an LSTM model to compare conventional catchment attributes and AEF embeddings as alternative static basin representations. The model is a 1D (input) to 1D (output) dynamical systems model, where the inputs are meteorological time series data spatially aggregated over the catchment, and the output is a streamflow time series at the corresponding gauge location. The model is trained using data from multiple catchments simultaneously. Because the objective of this study is to evaluate the relative utility of different static basin representations rather than to maximize predictive performance through extensive hyperparameter optimization, a fixed LSTM architecture was adopted across all experiments to ensure a fair comparison between basin descriptors. A grid search was conducted to select the key hyperparameters. The combination of 128 hidden states and a dropout rate of 0.4 achieved the best validation performance and was therefore adopted in all experiments. The resulting model consists of a single LSTM layer with 128 hidden states, a dropout rate of 0.4, and a linear output layer, trained using a batch size of 256. For prediction, the model parameters from the epoch achieving the best validation NSE were retained. The model was trained over the period from January 1980 to December 2004, validated from January 2005 to December 2009, and evaluated over the test period from January 2010 to December 2014, using an input sequence length of 365 days. Additional implementation details, including optimization settings and training configurations, are provided in Supporting Information Text S3.

We ran two different variants thereof. We trained two distinct models, each incorporating different sets of watershed attributes (the CAMELS attributes and AEF embeddings) which were concatenated with dynamic meteorological inputs. Each model variant was evaluated under both in-sample (IS) and out-of-sample (OOS) settings. Here, IS refers to temporal out-of-sample but spatial in-sample evaluation, whereas OOS refers to both temporal and spatial out-of-sample evaluation.

Model training and evaluation were conducted using five random seeds and a five-fold cross-validation scheme across the 671 CAMELS-US basins. In each fold, 80% of the basins were used for training, and the remaining 20% were held out as spatially independent test basins. For each fold and each random seed, the model was trained on the calibration period of the training basins and evaluated over the common test period from 2010 to 2014. The IS results were obtained from the test-period performance of the training basins, while the OOS results were obtained from the test-period performance of the held-out basins. Therefore, the IS setting assesses whether the model can generalize to unseen years for basins included during training, whereas the OOS setting assesses whether the model can generalize to both unseen years and unseen basins.

Model performance was evaluated using the Nash–Sutcliffe Efficiency (NSE, Nash and Sutcliffe, 1970) and the Kling–Gupta Efficiency (KGE, Gupta et al., 2009), both calculated over the test period. To further quantify uncertainty in the empirical performance distributions, we conducted a bootstrap analysis at the evaluation stage (Heudorfer et al., 2025). For each basin descriptor representation and each evaluation setting, the basin-level NSE values obtained across the five random seeds and five cross-validation folds were pooled into a single performance sample. Bootstrap resampling was then performed by repeatedly drawing 80% of the pooled basin-level NSE values with replacement.

In each bootstrap iteration, the empirical cumulative distribution function (CDF) was evaluated on a common predefined NSE grid, ensuring consistent comparison across bootstrap realizations. This procedure was repeated 5000 times, producing an ensemble of bootstrap CDF realizations for each model configuration. The final CDF curve was obtained by taking the median CDF value across all bootstrap realizations at each NSE grid point, and the 5th–95th percentile range was used to define the uncertainty envelope. Compared with single-point summary statistics, this distribution-based approach explicitly captures performance variability associated with random initialization, cross-validation basin splits, and finite-sample evaluation, thereby providing a more robust basis for comparing different basin descriptor representations under both IS and OOS settings.

In addition, distributional differences between model configurations were evaluated using two-sided Kolmogorov–Smirnov (KS) tests, assuming as the null hypothesis that the compared score distributions were identical (Heudorfer et al., 2025; Hodges, 1958). As a non-parametric test, the KS test makes no assumptions about the underlying distribution of performance scores, making it well suited for assessing whether observed differences between model configurations are statistically significant (Virtanen et al., 2020).

To further investigate the degree to which AEF embeddings encode complementary or redundant information relative to the traditional CAMELS attributes, we conducted a mutual information (MI) analysis between the two sets of static descriptors. Mutual information is an information-theoretic measure that quantifies the amount of shared information between two random variables and reflects the strength of their statistical dependence regardless of whether the relationship is linear or nonlinear. In this study, each CAMELS attribute and each AEF embedding dimension was treated as an independent random variable, and all possible pairs of attributes and embedding dimensions were compared to construct a complete mutual information matrix. The MI values allow us to determine whether specific AEF dimensions share substantial structure with known physiographic, climatic, or geological attributes or instead capture information that is not well represented in existing CAMELS descriptors.

For two discrete variables $X$ and $Y$, mutual information is formally defined as

$$I(X;Y) = \sum_{x \in \mathcal{X}} \sum_{y \in \mathcal{Y}} p(x,y) \log \frac{p(x,y)}{p(x)p(y)}$$

where $p(x,y)$ is the empirical joint probability and $p(x)$ and $p(y)$ are the corresponding marginal distributions. All variables were standardized to ensure comparability across features and to reduce the effects of scale differences.

The resulting MI matrix serves as a diagnostic tool for evaluating information redundancy between the two descriptor sets. High MI values reveal embedding dimensions that closely align with known watershed characteristics, whereas consistently low MI values suggest that AEF representations encode distinct and potentially hydrologically meaningful patterns not captured by traditional CAMELS attributes. This analysis complements the simulation experiments by offering a quantitative assessment of how much novel information AEF embeddings contribute beyond existing static attributes.

### Experiment B

The purpose of Experiment B is to examine whether choosing basins that share similar hydrological characteristics, instead of relying on the full set of available basins, can lead to improved prediction performance in ungauged regions. A further aim is to understand whether different definitions of basin similarity influence how well the model generalizes as the size of the training set gradually increases. We further examine the effects of donor-basin selection and static catchment representations by systematically combining different donor-selection strategies with both types of static inputs.

To conduct this analysis, we rank all basins in the CAMELS dataset with respect to a given target basin using two similarity measures defined in Section 2.3.

Based on these rankings, we construct seven training sets by selecting the top k most similar basins, where $k \in \{100, 200, 300, 400, 500, 600\}$, as well as a baseline using all 670 basins except the target basin. We also include randomly selected donor basins as a baseline for comparison. Three donor-selection strategies are combined with two static input representations (CAMELS attributes and AEF embeddings), resulting in six experimental configurations. Each training set is used to train the same LSTM model employed in Experiment A, and performance is evaluated on the target basin using NSE. This design allows us to systematically study how prediction skill scales as the training set expands from strongly similar basins to increasingly dissimilar ones.

The analysis was conducted on 100 target basins selected from the CAMELS-US dataset through stratified sampling based on the Köppen Climate Classification system (Beck et al., 2018), representing approximately 15% of the 671 basins. The centroid of each CAMELS basin was spatially overlaid with the North American climate-zone dataset to determine its corresponding Köppen climate subtype. The basins were then grouped into five major hydro-climatic regions: temperate, arid, semi-arid, continental, and alpine climates. To avoid overrepresentation of dominant climate types and ensure coverage across diverse hydrological regimes, at least one basin was sampled from each available Köppen climate subtype, while larger subtypes were allowed to contribute multiple basins. The spatial distribution of the 100 selected target basins, classified by major climate zone and Köppen climate subtype, is shown in Figure 3.

Experiment B covered 100 target basins, seven training-basin pool sizes, and six experimental configurations combining three donor-selection strategies with two static input representations. The experiment was designed to systematically assess how training-basin pool size affects model performance and to test whether increasing the number of training basins necessarily leads to continued improvements in predictive skill. We further summarized the distribution of donor-pool sizes associated with the highest NSE for each target basin and experimental configuration.

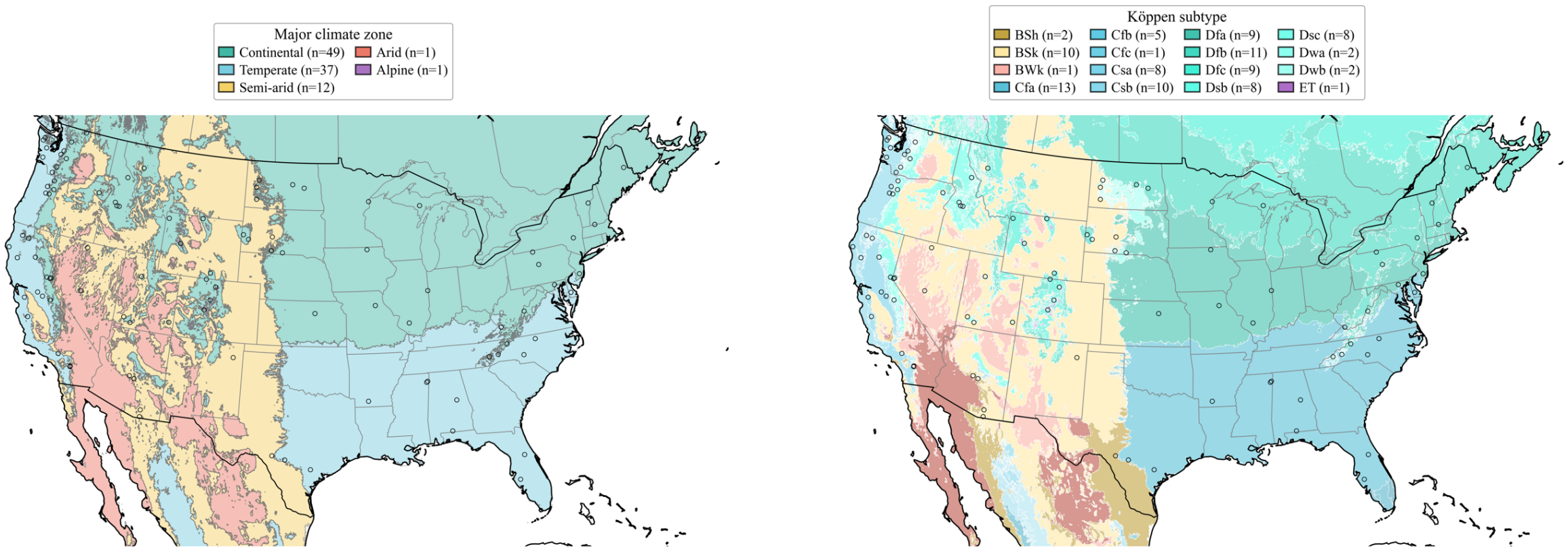

(a) Major climate zones (b) Köppen climate subtypes

Figure 3. Spatial distribution of the 100 selected target basins by major climate zone and Köppen climate subtype. Panel (a) shows the distribution of selected basins across the five major climate zones, while panel (b) shows their corresponding Köppen climate subtypes. Each point represents a selected target basin, with colors indicating the assigned climate category. The numbers in parentheses in the legends denote the number of target basins sampled from each major climate zone or Köppen climate subtype.

# 4 Results

## 4.1 The Value of AEF Embeddings in Hydrological Simulations

In this experiment, AEF embeddings and CAMELS attributes were separately combined with meteorological forcings as inputs to LSTM models, and their performance was compared under IS and OOS settings. The IS setting evaluates temporal generalization for basins used during training, whereas the OOS setting evaluates generalization to spatially unseen basins.

The results indicate that the benefit of AEF embeddings is limited under the IS setting but becomes more evident under the OOS setting. As shown in Figure 4, the NSE distributions of the CAMELS-attribute model and the AEF-embedding model are nearly indistinguishable under the IS scenario (KS = 0.0177, p-value = 0.0779). In contrast, under the OOS scenario, the two models exhibit a modest but statistically significant difference in their performance distributions (KS = 0.0863, p-value = $6.07 \times 10^{-9}$), indicating that AEF embeddings provide greater value when models are transferred to spatially unseen basins. A consistent pattern was observed for Kling–Gupta Efficiency (KGE), as shown in Figure S2.

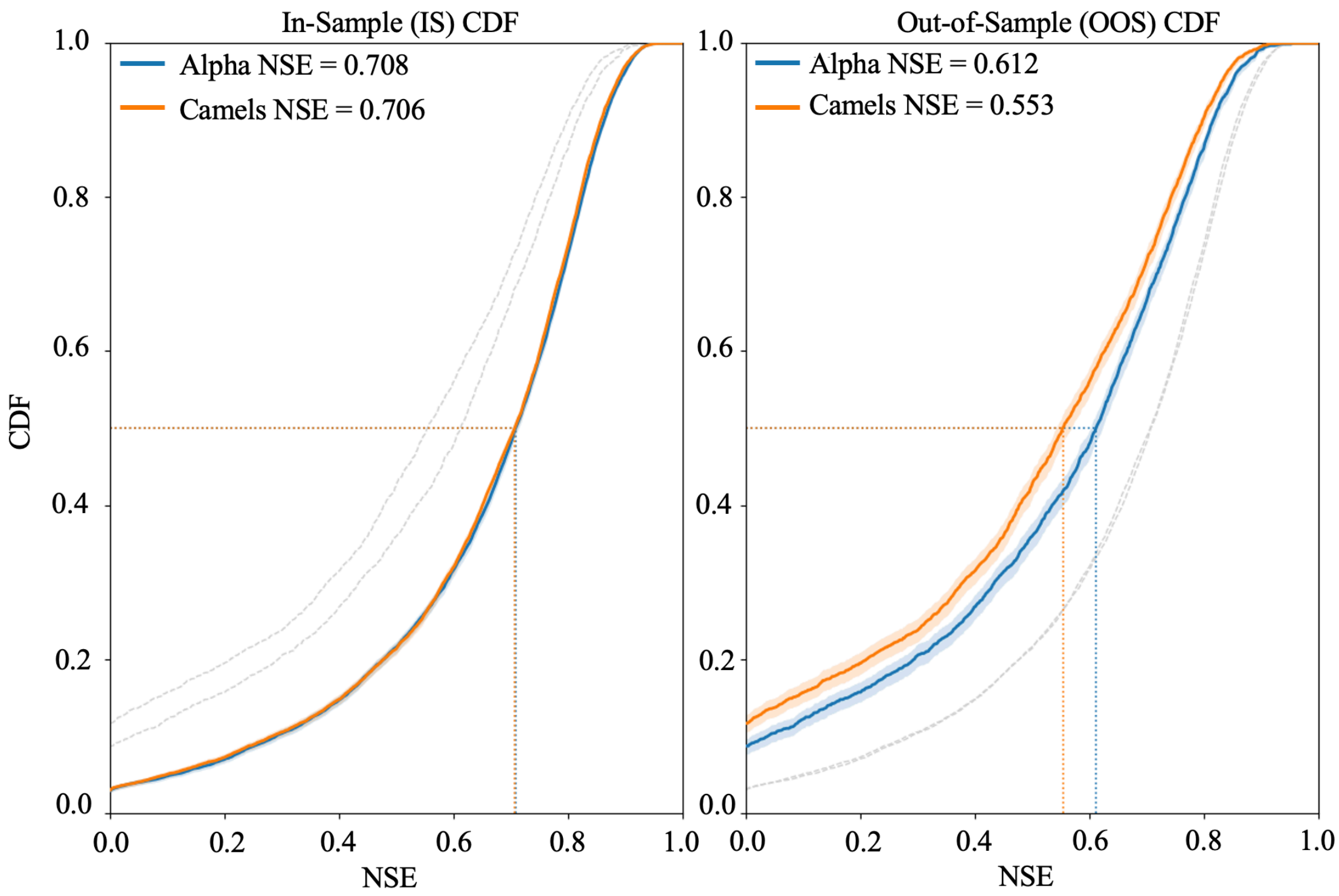

Figure 4. Cumulative distribution function (CDF) of the Nash-Sutcliffe Efficiency (NSE) for two different models in spatial in-sample (IS) and out-of-sample (OOS) settings. Solid curves represent the median bootstrap CDF obtained from 5000 bootstrap realizations, and shaded regions indicate the corresponding 5th–95th percentile uncertainty envelopes.

Although both models experience a performance decline when moving from IS to OOS, the reduction is substantially larger for the CAMELS-attribute-based model. This indicates that the model using conventional basin attributes exhibits a greater decline in predictive performance on spatially unseen basins. In contrast, the AEF-embedding-based model exhibits a considerably smaller degradation, suggesting that the AEF embedding provide a more transferable representation of basin characteristics and support improved predictive generalization under distributional shifts. Most importantly, the AEF-embedding-based model outperforms the CAMELS-attribute-based model under OOS conditions. This suggests that AEF encapsulates richer and more informative representations of basin characteristics, enabling the model to extract underlying relationships that are more relevant for streamflow simulation than those captured by traditional static attributes.

To examine whether the predictive advantage of AEF embeddings could be explained by their dimensionality alone, we conducted additional experiments using fixed 64-dimensional random vectors as basin descriptors. While the random-vector model achieved comparable performance under IS conditions, its median NSE dropped to 0.137 under OOS conditions, substantially below that of AEF embeddings (0.612). These results suggest that the predictive value of AEF embeddings cannot be attributed solely to their dimensionality (Supporting Information Text S5 and Figure S3).

Further evidence of the environmental characteristics associated with AEF embeddings is provided by the mutual information analysis between the CAMELS attributes and the AEF embedding (Figure 5). The results show that several CAMELS terrain and vegetation attributes exhibit substantial shared information with the AEF space, including elevation, slope, LAI metrics, and dominant land cover. This pattern is expected given that topography and vegetation can be directly inferred from remote-sensing imagery, and root depth is strongly linked to vegetation type and thus partly recoverable through indirect cues. Many other CAMELS attributes, however, show relatively low pairwise mutual information with individual AEF dimensions. These weaker associations may reflect differences in how catchment information is encoded. AEF embeddings may also capture aspects of land-surface and ecohydrological dynamics not explicitly represented in CAMELS, which could contribute to their better OOS performance.

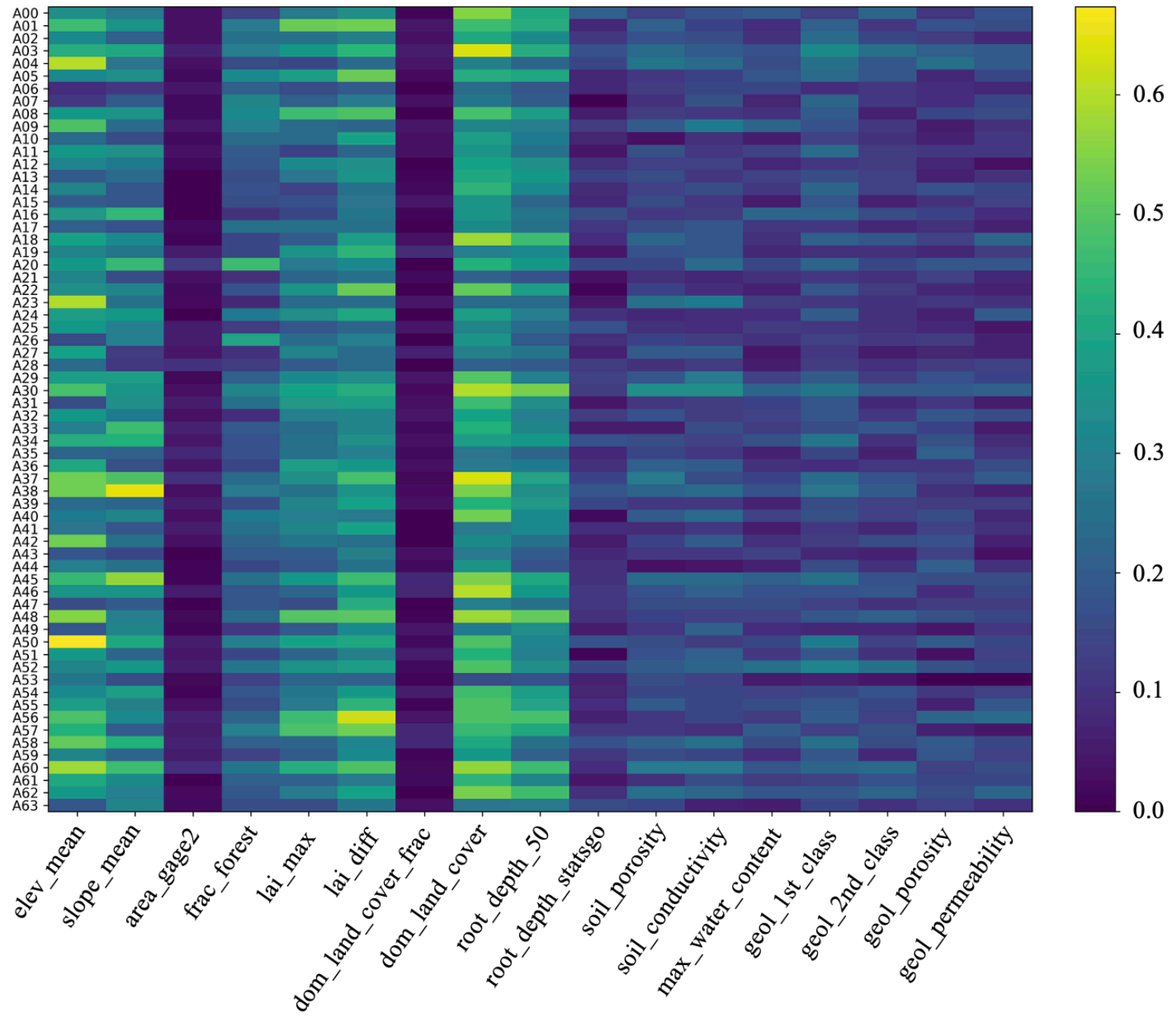

Figure 5. Mutual information matrix between the 64-dimensional AEF embedding and the CAMELS static attributes.

The relative robustness of the AEF model raises questions about how deep learning models leverage physiographic attributes. Our results suggest that static attributes may partially function as identifiers for basins in large-sample learning systems, enabling models to learn basin-specific relationships under in-sample conditions that do not necessarily generalize to unseen basins (Li et al., 2022; Heudorfer et al., 2025). Conversely, being derived from large-scale, self-supervised learning across millions of satellite-grid samples, AEF embeddings may encode broader environmental patterns that support predictive generalization across basins. These patterns may be relevant to hydrological functioning, although the specific environmental information responsible for the observed predictive advantage remains unclear.

In summary, these results demonstrate that AEF embeddings improve predictive performance in spatially unseen basins compared with traditional CAMELS attributes under the evaluated experimental settings, raising an important question: how can such representations be most effectively used to improve cross-basin generalization?

## 4.2 The Scaling Effect for Using AEF Embeddings

Building on the findings in Section 4.1 that AEF embeddings improve spatial out-of-sample prediction compared with CAMELS attributes, we further investigate how donor-pool size, donor-selection strategies, and static basin representations affect PUB performance, and whether smaller donor pools can achieve performance comparable to or better than larger ones.

To understand the mechanism driving potential performance differences, we first examine the topological structure of the similarity spaces defined by the two methods (Figure 6). The pairwise similarity matrices in Figure 6(a) reveal distinct patterns between the two representations. The CAMELS attribute-based similarity matrix exhibits a relatively fragmented structure, with substantial variation in similarity across basin pairs. In contrast, the AEF-based matrix displays a more coherent structure, characterized by generally higher similarity values and more clearly defined groups of basins with similar embeddings. These differences indicate that CAMELS attributes and AEF embeddings characterize basin relationships differently within their respective representation spaces.

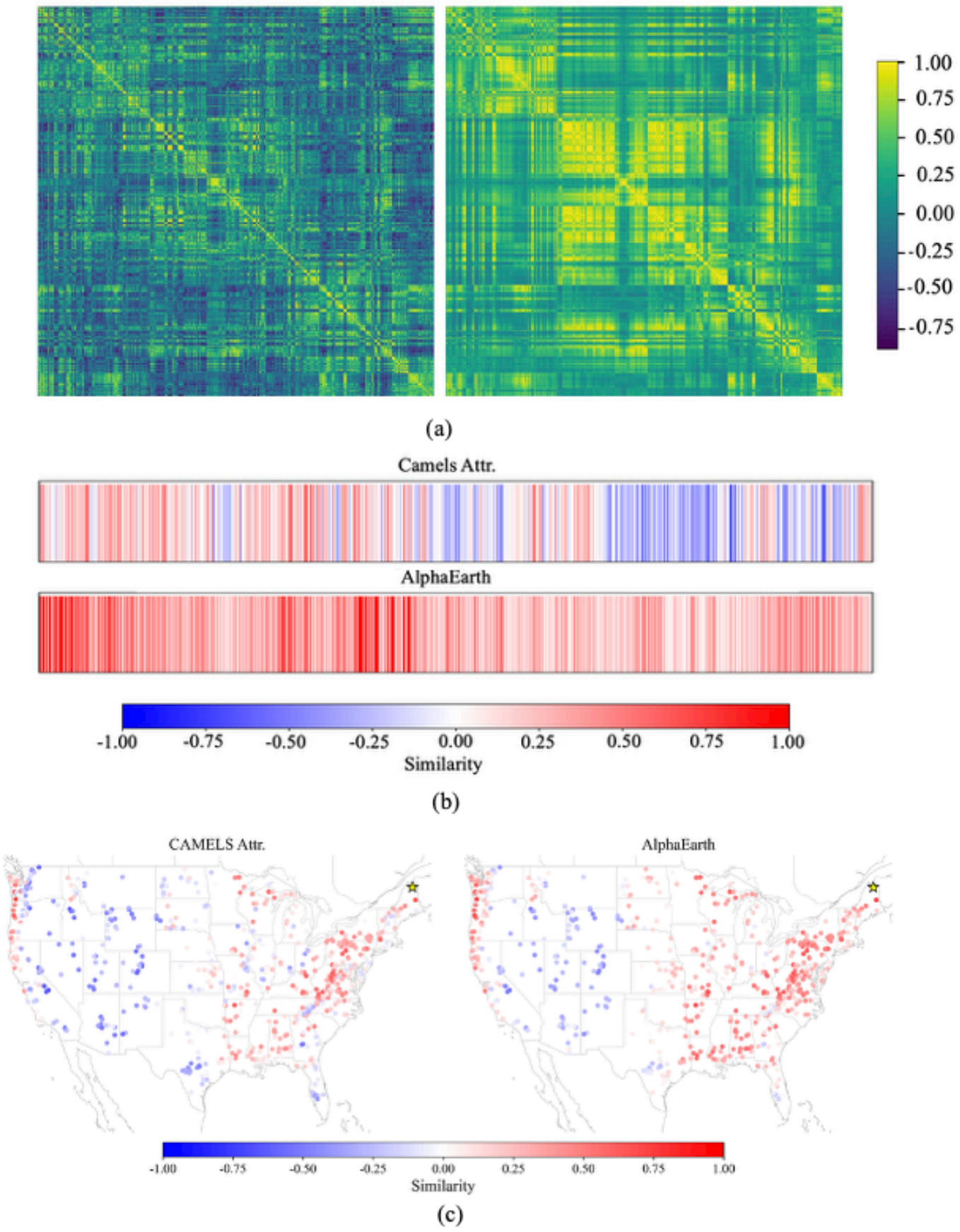

Figure 6. Comparison of basin similarity patterns derived from CAMELS attributes and AlphaEarth Foundation (AEF) embeddings. (a) Pairwise cosine similarity matrices among CAMELS-US basins under the two representations. (b) Similarity stripe plots for the selected target basin 01013500, where each vertical stripe represents a candidate basin and its color indicates the cosine similarity to the target. (c) Spatial distributions of basin similarity for the same target basin, with each point representing a basin and the star marking the target basin.

To visualize these patterns at the level of a single basin, Figure 6(b) presents stripe plots for the target basin 01013500. Each vertical stripe represents one candidate basin, and the color encodes its similarity to the target under each method. Under CAMELS attributes, stripes fluctuate irregularly, showing scattered pockets of high similarity interspersed with low-similarity basins. Under AEF embeddings, however, we observe a tighter, more coherent band of high similarity, with fewer abrupt fluctuations. In addition, Figure 6(c) presents the corresponding spatial distribution maps for the same target basin. Under CAMELS attributes, basin similarity patterns appear spatially fragmented and heterogeneous, with scattered regions of inconsistent similarity. In contrast, AEF embeddings generate a smoother and more spatially continuous similarity field, with broadly elevated similarity values across hydrologically related regions and fewer isolated low-similarity outliers. These results suggest that AEF embeddings capture distinct spatial patterns of basin similarity across the CAMELS-US domain.

A central question in Experiment B is how PUB performance changes as the donor pool expands, and how this scaling behavior depends on donor selection and static catchment representation. As shown in Figure 7(a), we first compare the overall performance of the six experimental configurations across different training-basin pool sizes. Configurations using AEF embeddings as static inputs generally achieve higher median NSE than those using CAMELS attributes, while the effectiveness of different donor-selection strategies varies with donor-pool size.

Across the evaluated donor-pool sizes, configurations using AEF embeddings as static inputs generally achieve higher median NSE than those using CAMELS attributes, with an average NSE advantage of 0.371 across the seven donor-pool sizes and three donor-selection strategies. With AEF inputs, non-random donor selection generally outperforms random selection for smaller donor pools, but this advantage diminishes as the pool expands. With CAMELS inputs, the benefits of similarity-based selection are less consistent and vary with the selection rule and donor-pool size. These differences are more evident in predictions for ungauged target basins than in test-period predictions for donor basins included during training, where the AEF-based and CAMELS-attribute models perform broadly similarly, consistent with the IS results of Experiment A (Supporting Information Text S6 and Figure S5). These results support the value of AEF embeddings for transferring predictive relationships from donor basins to ungauged target basins.

Although static input representation appears to have a substantial influence on PUB performance, donor selection also affects the scaling behavior, particularly for smaller donor pools. With AEF inputs, CAMELS-based and AEF-based donor selection yield broadly comparable predictive performance and generally outperform random selection at smaller pool sizes. As shown in Figure 7(b), the two similarity-based strategies select donors with increasingly similar AEF-space similarities to the target basins as the donor-pool size increases, with the median difference decreasing from 0.185 at $k = 100$ to 0.022 at $k = 600$ and substantial overlap in their 25th–75th percentile ranges. Both strategies generally yield higher similarities than random selection. This may help explain their comparable predictive performance and their advantage over random selection for smaller donor pools. As the donor pool expands, these similarity differences decrease, accompanied by a diminishing predictive advantage over random selection. With CAMELS inputs, the benefits of similarity-based selection are less consistent and vary with the selection rule and donor-pool size. Multi-seed experiments on 10 target basins also show greater seed-to-seed variability in the CAMELS-based configuration for several basins, indicating that some performance differences are sensitive to random initialization (Supporting Information Text S7 and Figure S6).

However, the trend shown in Figure 7(a) represents an average pattern across the 100 target basins and should not be interpreted as evidence that larger training-basin pools are always preferable. Figure 7(c) summarizes the distribution of the optimal pool size, defined as the value of k that produces the highest NSE for each target basin and experimental configuration. These optimal pool sizes were determined retrospectively using observed target-basin streamflow. The results show substantial basin-to-basin variability. For the configuration using AEF-based donor selection with AEF inputs, only 16% of the target basins achieve their highest NSE when all 670 donor basins are used, whereas approximately 60% reach their best performance with $k \leq 400$. This indicates that many basins achieve their highest predictive performance with smaller donors rather than the largest possible training pool. Similar variability is also observed across the other experimental configurations, suggesting that no single training-basin pool size is universally optimal across all target basins. It should be noted that this is a retrospective analysis using observed target-basin streamflow; the identified optima may vary with random initialization and cannot serve as practical pool-size selection rules for ungauged basins.

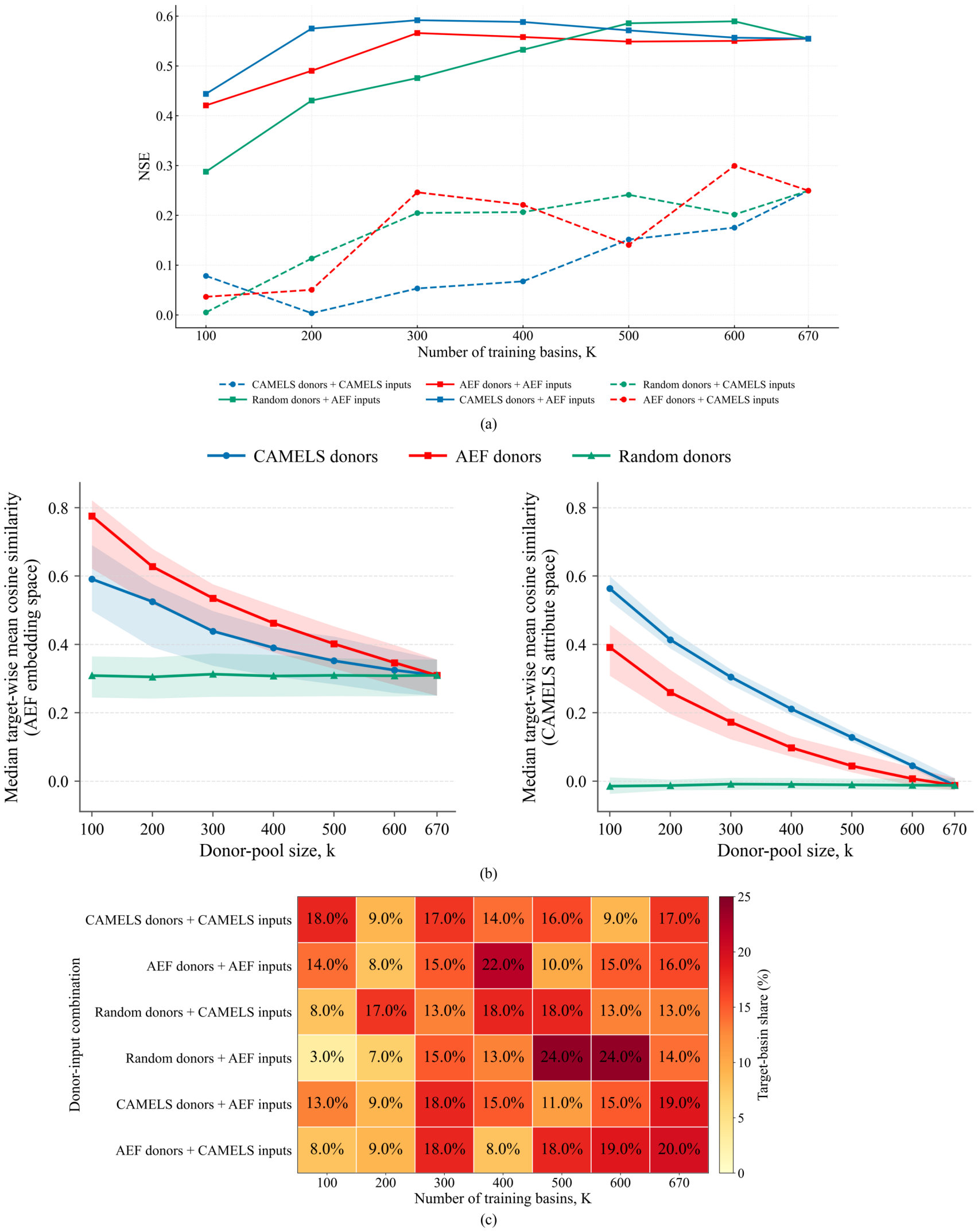

Figure 7. Effects of training-basin pool size on PUB performance across 100 target basins under six experimental configurations combining similarity-based or random donor selection with CAMELS attributes or AEF embeddings as static inputs. Panel (a) shows the median NSE across different training-basin pool sizes. Panel (b) shows the median cosine similarity between target basins and selected donor basins in the CAMELS attribute space and AEF embedding space under three donor-selection strategies, with shaded areas indicating the 25th–75th percentile range across target basins. Panel (c) shows the distribution of retrospectively optimal training-basin pool sizes, determined using observed target-basin streamflow to identify the highest NSE for each target basin.

This basin-level heterogeneity becomes clearer in Figure 8, which presents NSE curves for five selected target basins from the continental, temperate, alpine, semi-arid, and arid climate zones. The curves show that responses to increasing donor-pool size differ substantially among basins and among experimental configurations. For some basins, relatively small or moderate donor pools already achieve competitive or superior performance, whereas for others, performance continues to improve as more basins are included. The selected arid-climate basin exhibits relatively poor predictive performance across the evaluated configurations. These examples reinforce the interpretation from Figure 7(c) that PUB performance does not follow a simple "more basins are always better" scaling rule. Instead, the donor-pool size associated with the highest NSE varies across target basins and depends on the static basin representation and donor-selection strategy used.

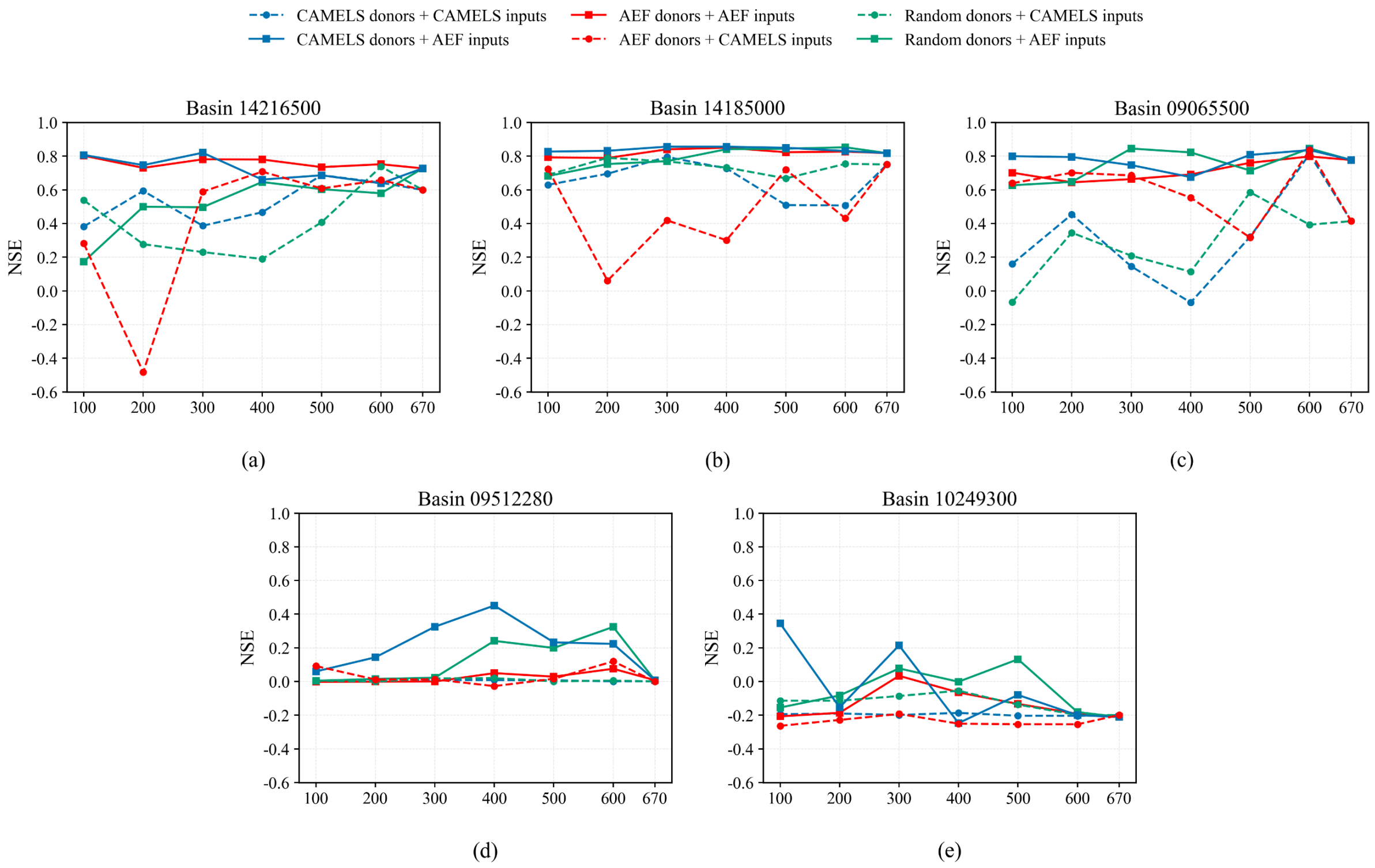

Figure 8. NSE performance under different training-basin pool sizes for five target basins randomly selected from the continental, temperate, alpine, semi-arid, and arid climate zones, using six experimental configurations combining three donor-selection strategies (CAMELS attributes, AEF embeddings, and random selection) with two static input representations (CAMELS attributes and AEF embeddings). Panels (a)–(e) correspond to the continental, temperate, alpine, semi-arid, and arid climate zones, respectively. Each panel presents the NSE curve for an individual basin.

Overall, these results indicate that static basin representation plays a more important role in PUB performance than the choice of donor-selection strategy under the evaluated settings. Configurations using AEF embeddings generally achieve higher predictive performance across different donor-selection strategies, including random selection. Donor selection and pool size nevertheless influence the results, with the benefits of similarity-based selection being more apparent for smaller donor pools when AEF inputs are used. Larger donor pools do not necessarily lead to better predictive performance, and the benefits of similarity-based donor selection vary across donor-pool sizes and static input representations.

## 5 Discussion

A central goal of this study was to evaluate whether Earth foundation model embeddings, specifically the 64-dimensional AEF embeddings, can improve generalization in rainfall-runoff simulation. Our results support this potential and highlight the value of AEF embeddings as static model inputs, while showing that the benefits of donor-basin selection depend on training-set size. However, because these representations are learned from large-scale satellite observations without hydrological supervision, the catchment information they encode warrants further examination. Furthermore, it is important to investigate how the predictive benefits of AEF embeddings vary when models are evaluated on basin groups defined by different static representations, particularly when entire groups are excluded from training.

### 5.1 Catchment Organization in the AEF Representation Space

To better understand how AEF embeddings characterize differences among catchments, we compare their organization of basins with that derived from conventional CAMELS attributes. Specifically, we apply k-means clustering to the 671 CAMELS-US basins based on either the CAMELS attributes or the AEF embeddings, assigning each basin to one of ten clusters (C0–C9). This setting controls for differences in cluster numbers and provides a consistent basis for the subsequent comparison of basin groupings and cross-regime prediction. We then compare the spatial distributions of the two partitions.

Figure 9 shows the spatial distributions of the ten basin groups derived from CAMELS attributes and AEF embeddings. These distributions illustrate how the two representations organize the CAMELS-US basins into different groups. The AEF-derived groups appear more spatially coherent than those derived from CAMELS attributes, with fewer changes in cluster membership among neighboring basins. However, this spatial coherence should not be equated with greater hydrological similarity or easier predictive transfer. In Section 4.2, AEF embeddings improved predictions as static inputs, but AEF-based donor selection did not consistently outperform CAMELS-based selection. The predictive value of this catchment organization may therefore depend on how the model uses the representation.

Despite their predictive value, it remains difficult to determine what information individual AEF dimensions represent. AEF embeddings provide a compact latent representation of complex, multi-source observations. As described in Section 2.2, one of the training objectives is to reconstruct source observations from these embeddings using observation times and sensor metadata. This objective encourages the embeddings to retain information useful for reconstruction. However, it does not require individual dimensions to correspond to named catchment properties or hydrological processes. The associations identified in Section 4.1 and spatial groupings examined here provide clues to the information represented, rather than a direct interpretation of each dimension (Supporting Information Text S8 and Figure S8).

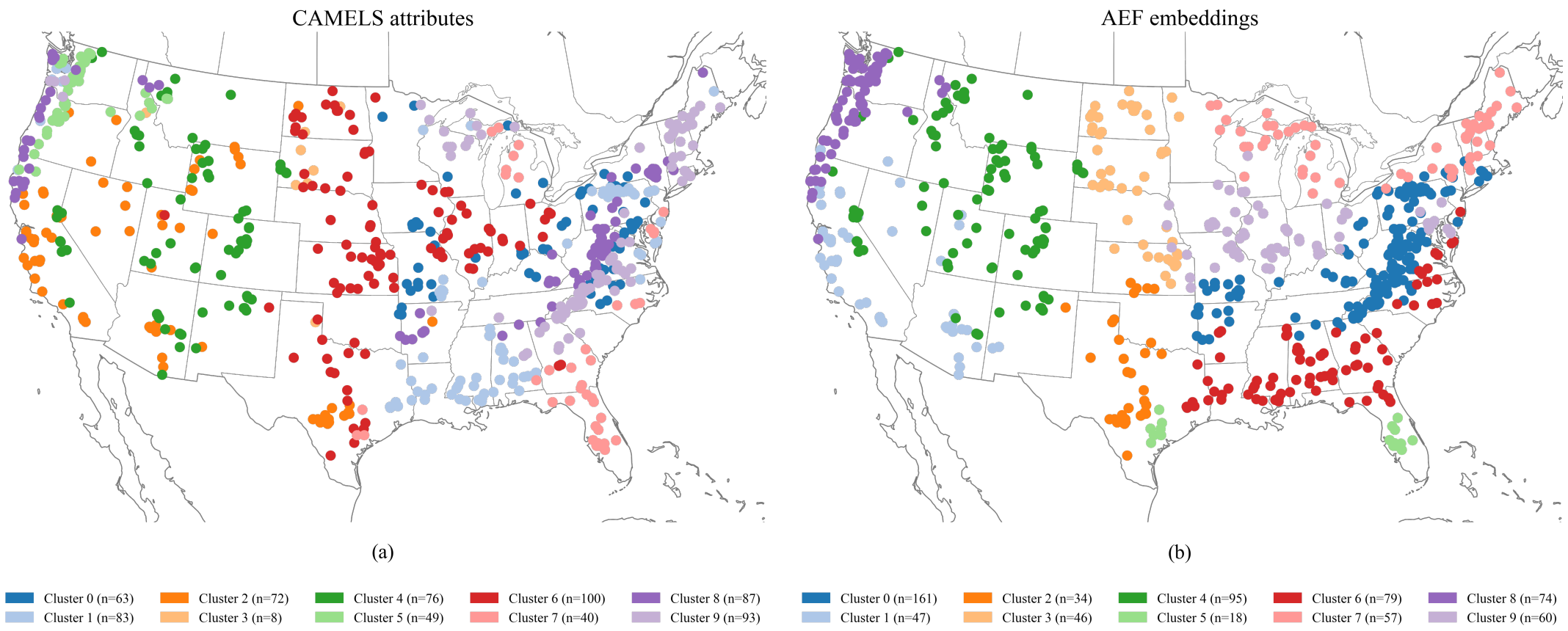

Figure 9. Spatial distribution of basin clusters derived from (a) CAMELS attributes and (b) AEF embeddings. Basins are partitioned into 10 clusters under each representation, with different colors indicating cluster membership. The maps illustrate differences in the spatial clustering patterns of CAMELS-US basins between the two representations.

## 5.2 Limits of Extrapolation: Cross-Regime Generalization Analysis

To further evaluate model transferability across the basin groups examined in Section 5.1, we conducted a cluster-based cross-validation experiment. Instead of using fixed geographical regions like ecoregions (Feng et al., 2021; Omernik and Griffith, 2014), we defined "regimes" by clustering basins based on their feature spaces (CAMELS attributes vs. AEF embeddings). For each representation, we used the same ten-cluster partition as in Section 5.1. We evaluated both CAMELS attributes and AEF embeddings as static model inputs under each clustering scheme, resulting in four experimental configurations, each repeated with five random seeds. Because cluster sizes differ, the number of training basins varies depending on which cluster is withheld for testing, and the resulting transfer tasks may differ in difficulty.

Figure 10 summarizes the aggregated performance across all regimes (all 671 basins). Under CAMELS-based clustering, AEF inputs achieved a median NSE of 0.477, compared with 0.374 for CAMELS inputs. Under AEF-based clustering, the corresponding median NSE values were 0.349 and 0.355, indicating similar performance for the two input representations. While the difference between clustering schemes was relatively small when CAMELS attributes were used as static inputs, AEF inputs achieved substantially higher median NSE under CAMELS-based clustering than under AEF-based clustering. These results suggest that the predictive benefits of AEF embeddings observed in Section 4.2 do not necessarily extend to settings in which entire AEF-defined basin groups are excluded from training.

This contrast may reflect the dependence of cross-regime transferability on the catchment conditions represented during training. AEF embeddings provide informative descriptions of basins, which is beneficial when similar regimes are represented in the training data. However, withholding an entire AEF-defined cluster may reduce the coverage of similar catchment conditions, potentially limiting the benefits of AEF inputs. In other words, AEF embeddings can support cross-basin prediction, as shown in Section 4.2, but this advantage becomes harder to exploit when the target regime is poorly represented in the training basins. In contrast, under CAMELS-based clustering, the resulting clusters are less distinct, meaning that the "training" clusters likely still contain basins that broadly resemble the "test" cluster with AEF inputs. Moreover, the two clustering schemes yielded similar median performance when CAMELS attributes were used as inputs. A more informative static representation may therefore improve predictions without ensuring that the model learns rainfall–runoff relationships that transfer across distinct basin groups.

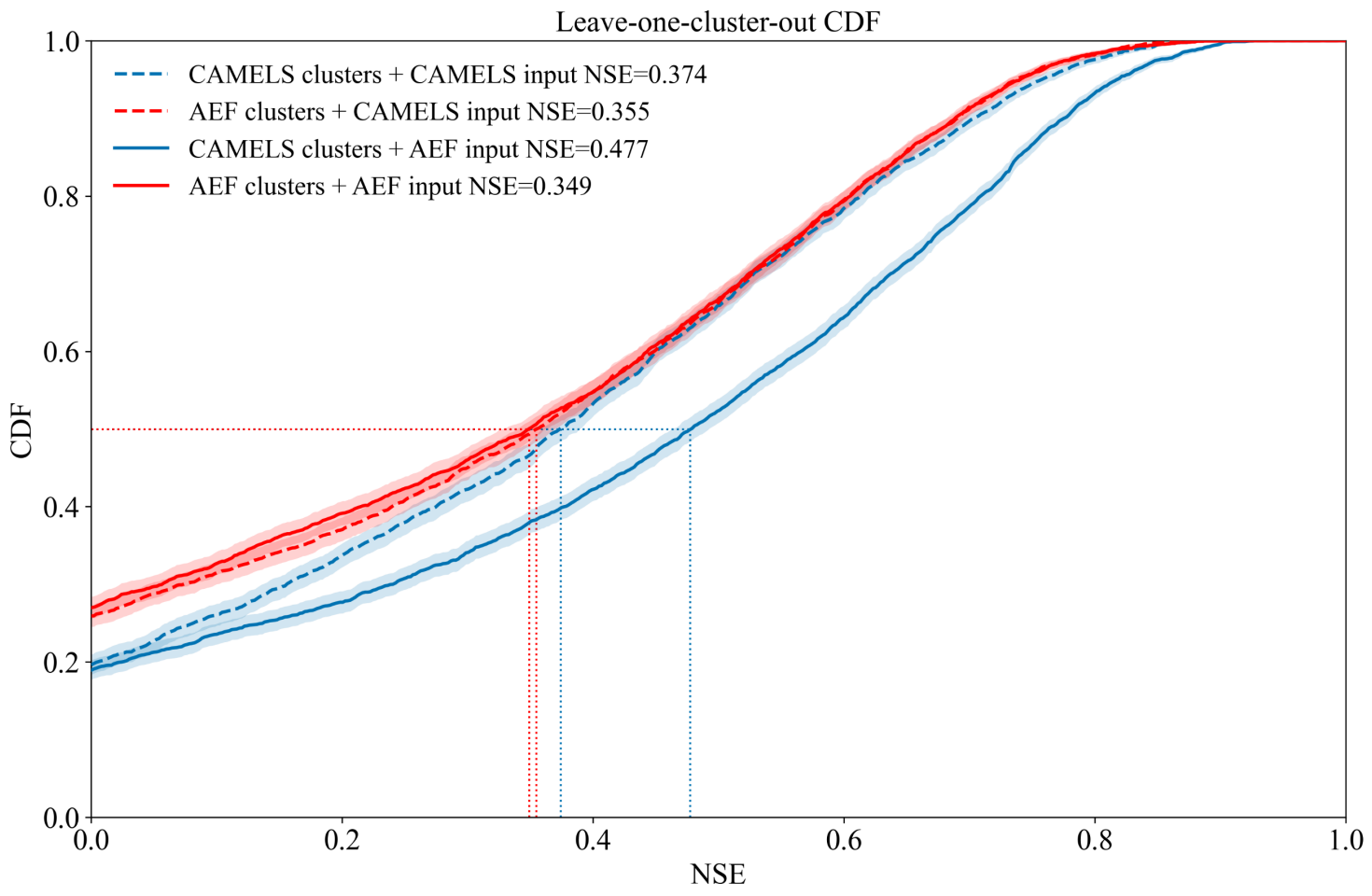

Figure 10. Cumulative distribution functions (CDFs) of Nash–Sutcliffe efficiency (NSE) from the leave-one-cluster-out evaluation under four configurations combining CAMELS- or AEF-based clustering with CAMELS or AEF static inputs. For each configuration, results from five random seeds were pooled, and 5,000 bootstrap replicates were generated by independently sampling 80% of the basin–seed observations with replacement. At each NSE threshold, the solid line represents the median cumulative probability across bootstrap replicates, while the shaded band is bounded by the corresponding 5th and 95th percentiles. The horizontal dotted line indicates a cumulative probability of 0.5, while the colored vertical dotted lines mark the median NSE values.

This finding has broader implications for the pursuit of a unified hydrological model. The fact that performance with AEF inputs differs substantially between the two clustering schemes suggests that, given current data constraints, a single global model may struggle to optimally represent the full diversity of hydrological behaviors. Although AEF embeddings provide informative catchment descriptions, supplying them as static inputs to a supervised rainfall–runoff model may not be sufficient to learn transferable representations of catchment hydrological behavior.

Therefore, our results suggest that improving static catchment descriptions alone does not guarantee reliable transfer across distinct catchment conditions. Moving towards a truly unified model will likely require more than just embeddings from geospatial foundation models; it calls for exploring how models internally represent rainfall–runoff processes and learn relationships that remain applicable across different catchment conditions, alongside integrating more information from diverse data sources (Liu et al., 2026) and variables (Ouyang et al., 2025) to help the model separate general physical patterns from unique local characteristics.

## 6 Conclusions

This study systematically evaluated the predictive utility and transferability of AlphaEarth Foundations (AEF) embeddings as a new form of static basin representation for large-sample hydrological modeling and further examined their potential in prediction in ungauged basins (PUB). The results show that AEF embeddings provide stronger cross-basin generalization than traditional CAMELS attributes. Although both types of static features performed similarly in-sample, with median NSE values of 0.708 for the AEF embeddings and 0.706 for the model using CAMELS attributes, their behavior differed noticeably once they were evaluated in spatial out-of-sample settings. For these more challenging conditions, the median NSE of the AEF embeddings was 0.612, while the model relying on CAMELS attributes reached 0.553. Consequently, models incorporating AEF embeddings were more robust and yielded higher predictive accuracy across basins. Mutual-information analysis further revealed statistical associations between AEF embeddings and conventional catchment attributes related to topography, vegetation, and other environmental characteristics. These findings support the potential of Earth foundation model embeddings as informative static basin representations for improving hydrological prediction across basins.

In the scaling PUB experiments involving 100 target basins, we examined how donor-basin selection, static input representations, and donor-pool size affect predictive performance. Configurations using AEF embeddings as static inputs generally achieved higher median NSE than those using CAMELS attributes, with an average NSE advantage of 0.371 across seven donor-pool sizes and three donor-selection strategies. When AEF embeddings were used as static inputs, similarity-based donor selection generally outperformed random selection for smaller donor pools, but this advantage diminished as the pool size increased, with random selection achieving comparable or slightly higher performance at larger pool sizes. Moreover, increasing the number of donor basins did not consistently improve predictive performance, with substantial variability among individual target basins. These results indicate that PUB performance depends on donor-pool size and composition as well as the static representation used by the model. The strong performance of AEF embeddings under both similarity-based and random donor selection suggests that their predictive value extends beyond explicit similarity-based donor identification.

Importantly, our cross-regime experiments highlight that the predictive benefits of AEF embeddings depend on how basins are grouped for training and evaluation. Under CAMELS-based clustering, AEF embeddings achieved higher predictive performance than conventional attributes, with median NSE values of 0.477 and 0.374, respectively, whereas this advantage was not observed when entire AEF-defined clusters were withheld from training (0.349 versus 0.355). These findings suggest that AEF embeddings provide valuable catchment information, but their use as static inputs does not necessarily ensure consistent transfer across different basin groups. Whether global-scale datasets covering North America, Europe, East Asia, Australia, and beyond might reduce this dependence on the catchment conditions represented during training and enable models to learn truly transferable hydrologic patterns remains an open question requiring further investigation.

Beyond expanding data coverage, developing a unified hydrological AI model also requires examining how models internally represent catchment hydrological processes. Current rainfall–runoff modeling paradigms remain largely centered on learning statistical input–output relationships, rather than understanding the internal hydrological functioning of catchments. Although deep learning models such as LSTMs have demonstrated strong predictive performance under large-sample settings, it remains unclear whether these models genuinely learn transferable hydrological laws or merely fit statistical mappings associated with specific basin conditions. Future hydrological foundation models may therefore need to evolve beyond input–output rainfall–runoff mappings toward multi-level representations capable of characterizing latent hydrological structures and internal basin dynamics. Inspired by the development of Earth Foundation Models, integrating multi-source spatiotemporal observations, self-supervised learning, and latent-variable representations may enable models to learn more stable and transferable hydrological structures, rather than relying primarily on surface-level correlations. Such a transition may represent a critical step toward achieving true cross-basin generalization in hydrological modeling.

Overall, this study highlights the potential of Earth Foundation Models embeddings for large-sample hydrological modeling. It clarifies their predictive value under different donor-selection and training conditions, while identifying limits to their transfer across basin groups. We hope that these findings will encourage further research into the fundamental question of whether and to what extent deep learning models can capture transferable aspects of catchment hydrological functioning, while also contributing to the continued development of PUB theory and methodology.

## Acknowledgments

This research was supported by the Fund of the National Natural Science Foundation of China (Grant No. 52309010, 52322901) and the National Key Research and Development Program of China (Grant No. 2023YFF0804900). This research was also supported by the AI for Science (AI4S) Program of Dalian University of Technology.

## Data Availability Statement

The code and data supporting this study are permanently archived in Zenodo (https://doi.org/10.5281/zenodo.21121620) and are also available in the GitHub repository (https://github.com/CylenLC/AlphaEarth4PUB). The implementation uses the open-source hydrodataset and torchhydro packages. The CAMELS dataset is available at https://ral.ucar.edu/solutions/products/camels. AEF embedding data is available at https://developers.google.com/earth-engine/datasets/catalog/GOOGLE_SATELLITE_EMBEDDING_V1_ANNUAL.

## Conflict of Interest

The authors declare there are no conflicts of interest for this manuscript.